\documentclass[final]{cvpr}
\usepackage[table]{xcolor}
\usepackage{booktabs,multirow, array, tabularx}
\usepackage{times}
\usepackage{epsfig}
\usepackage{graphicx}
\usepackage{amsmath}
\usepackage{amssymb}
\usepackage[pagebackref=true,breaklinks=true,colorlinks,bookmarks=false]{hyperref}
\def\cvprPaperID{****}
\def\confYear{CVPR 2021}
\begin{document}
\title{Optical Mouse: 3D Mouse Pose From Single-View Video}
\author{Bo Hu*,
Bryan Seybold*,
Shan Yang*,\\
David Ross,
Avneesh Sud\\
Google LLC\\
{\tt\small \{bhuroc,seybold,shangyang,dross,asud\}@google.com}
\and
Graham Ruby, Yi Liu \\
Calico Life Sciences LLC\\
{\tt\small \{graham,yiliu\}@calicolabs.com}
}

\maketitle

\begin{abstract}
We present a method to infer the 3D pose of mice, including the limbs and feet, from monocular videos. Many human clinical conditions and their corresponding animal models result in abnormal motion, and accurately measuring 3D motion at scale offers insights into health. The 3D poses improve classification of health-related attributes over 2D representations. The inferred poses are accurate enough to estimate stride length even when the feet are mostly occluded. This method could be applied as part of a continuous monitoring system to non-invasively measure animal health.
\end{abstract}
\section{Introduction}

Many human clinical conditions and the corresponding animal models result in abnormal motion~\cite{burn2013oxford}. Measuring motion is a requisite step in studying the health of these subjects. For animal subjects, researchers typically conduct measurements manually at high cost, limited resolution, and high stress for the animals. In this work, we present a low-cost, non-invasive, computer-vision based approach for continuously measuring the motion as 3D pose of laboratory mice.

To study animal models of movement disorders, such as Parkinson's disease or tremor, or even generally measure behavior, researchers rely on manual tools such as the rotarod, static horizontal bar, open field tests, or human scoring~\cite{deacon2013measuring,gould2009open}. Increasingly complex automated tools to study gait and locomotion are being developed~\cite{dorman2014comparison,xu2019gait}. Computer vision and machine learning are creating new measurement opportunities in home-cage environments for 2D tracking or behavior~\cite{bains2018assessing,jhuang2010automated,kabra2013jaaba,mathis2018deeplabcut,noldus2001ethovision,pereira2019fast,richardson2015power}. So far, only a few studies measure 3D motion at all and only at coarse resolution or number of joints or requiring multiple cameras~\cite{dunn2021geometric,hong2015automated,salem2019three,sheets2013quantitative,wiltschko2015mapping}. Nevertheless, these new measurement tools are offering compelling opportunities for new analyses~\cite{dunn2021geometric,johnson2016composing,liu2019towards,wiltschko2015mapping}.

In parallel, computer vision and machine learning are leading to great improvements in determining human 3D pose from images. Models for optimizing a kinematic model to fit image data~\cite{bregler1998tracking} are being paired with improvements in estimating 2D poses~\cite{cao2018openpose,newell2016stacked,wei2016convolutional}. By combining these methods with libraries of human shapes~\cite{loper2015smpl} and human poses, 3D human pose estimates can be grounded to real kinematic models and realistic motions~\cite{bogo2016keep,pavlakos2018learning,tung2017self}. Ongoing research is improving the spatial and temporally coherence~\cite{arnab2019exploiting,humanMotionKZFM19,zanfir2018monocular}.

This work adapts these techniques originally developed to infer 3D human pose to mice. We predict 2D keypoints for mice then optimize for the 3D pose subject to priors learned from data. Databases of human shapes, poses, 2D keypoints, and 3D keypoints are readily available, but none of these are available for mice. The lack of data presented unique challenges to accurately infer 3D poses. We overcome these challenges by collecting new data and adapting where needed. We validate our method by demonstrating the metric accuracy of the inferred 3D poses, the predictive accuracy of health related attributes, and the correlation with direct measurements of gait. In each case, the inferred 3D poses are useful, detailed measurements. 

\section{Methods}\label{sec:methods}

\subsection{Data Collection}
We collect three sets of video data: Continuous, Multiview, and Gait. The \textbf{Continuous} video data is 14 days from 32 cages each outfitted with a single camera (Vium). During the dark cycle, infrared illumination is used. 8 animals are one-year old, knockout mice on a c57b6 background; 8 are one-year old, heterozygous controls; 8 are one-year old, c57b6 mice; and 8 are two-month old, c57b6 mice. The knockout mice have a deletion that causes motor deficits, the biology of which will be revealed in future work. The knockout mice and heterozygous controls are littermates on a c57b6 background, but have been inbred for several generations. Each mouse has three attributes, age (either twelve or three months old), knockout (either a full knockout or not), and background (either a littermate with a knockout or a c57b6). The \textbf{Multiview} video data is 35 consecutive multiview frames of a single c57b6 mouse in a custom capture rig (described below). The \textbf{Gait} video data is of a single c57b6 mouse walking on a treadmill with cameras installed below with corresponding commercial analysis tools (DigiGait) with an additional camera mounted above (GoPro) that we use for analysis. Each video is recorded at 24 to 30 frames per second. All experiments are approved by an Institutional Animal Care and Use Committee.

\subsection{Mouse Pose Prediction}
\begin{figure}[h]
\includegraphics[width=\columnwidth]{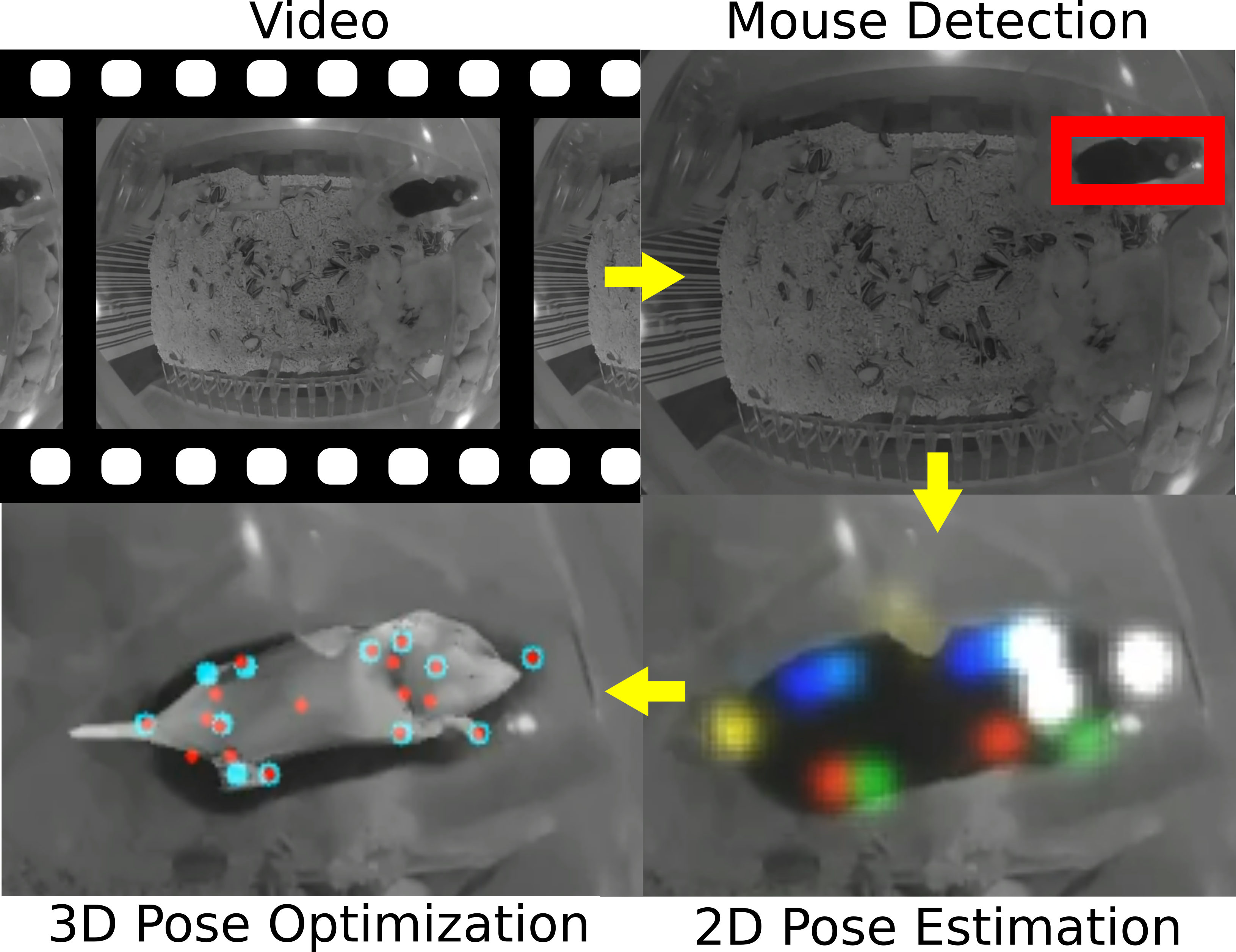}
\centering
\caption{Our pipeline operates over frames of a video (upper left). For each frame we run a 2D object detector trained to detect mice (upper right, box indicating a detection). We apply a 2D pose model to detect mouse key points at the detected location (lower right, colored heatmap indicating joint locations with arbitrary colors). Finally, we optimize for the 3D pose of the mouse (lower left, blue points are peaks of the keypoint heatmaps in previous stage, red points are projected 3D keypoints from the optimized pose, grey 3D mesh overlaid on the image).}
\label{feature_extraction_pipeline_fig}
\end{figure}

Our feature extraction pipeline (shown in Figure. \ref{feature_extraction_pipeline_fig}) includes three stages: bounding box detection, 2D pose prediction, and 3D pose optimization. These stages have been shown to be effective for human 3D pose estimation~\cite{bogo2016keep,lassner2017unite,varol2017learning}. 

\subsubsection{2D detection and pose prediction}
We adapt a Single-Shot Detector~\cite{liu2016ssd} to detect the mouse and a Stacked Hourglass Network~\cite{newell2016stacked} to infer the mouse's 2D pose similar to other work adapting human pose models to laboratory animals~\cite{mathis2018deeplabcut,pereira2019fast}. 

The detection and pose models both require training data, which we generated by labeling 20 joint positions along the body, and take the minimal box encompassing all points to be the bounding box. Models are pretrained on COCO~\cite{lin2014coco} and the prediction heads for human keypoints are replaced with those for mouse keypoints. For the Continuous video data, we label 3670 images for the training set and 628 for the test set. For the Gait video data, we fine-tune the Continuous video model on an additional 329 labeled image training set and test on 106 images. Frames are selectively annotated to cover the diversity of input images across cages and times.

We evaluate our pose model with the Object Keypoint Similarity (OKS) score used on COCO~\cite{lin2014coco}: $\sum_{i}\exp(-\mathbf{d}_i^2 / (2\mathbf{k}_i^2\mathbf{s}^2)) / 20$, where $\mathbf{d}_i$ is the Euclidean distance between the prediction and ground truth, $\mathbf{s}$ is the object scale as the square root of the bounding box area, and the per-keypoint falloff, $k_i$, is set to the human median of 0.08 for all keypoints\footnote{See http://cocodataset.org/\#keypoints-eval for further OKS details}. Accuracies are computed as the percentage of predicted keypoints greater than a threshold OKS score in Table~\ref{tab:2d_pose}.

\begin{table}
\centering
\begin{tabularx}{\columnwidth}{>{\centering\arraybackslash}X|*5{>{\centering\arraybackslash}X}@{}}
\toprule
    T & Nose & Shoulder & Hip  & Wrist & Ankle
    \\\midrule
     & \multicolumn{5}{c}{Continuous Video Test Dataset - Vium Cage}
\\\cmidrule(lr){2-6}
0.5 & 0.92 & 0.96 & 0.93 & 0.91 & 0.91\\
0.7 & 0.87 & 0.93 & 0.85 & 0.77 & 0.75\\
0.9 & 0.72 & 0.64 & 0.47 & 0.44 & 0.34\\
\midrule
 & \multicolumn{5}{c}{Gait Video Test Dataset - GoPro over Digigait}
\\\cmidrule(lr){2-6}
0.5 & 1. & 0.96 & 0.89 & 0.6 & 0.80\\
0.7 & 0.99 & 0.79 & 0.70 & 0.24 & 0.67\\
0.9 & 0.70 & 0.41 & 0.23 & 0.09 & 0.26     
\\\bottomrule
\end{tabularx}
\caption{The 2D pose accuracy as proportion of keypoints with OKS scores above the specified thresholds, T, for different joints across different data sets.}
\label{tab:2d_pose}
\end{table}

\begin{figure}[h]
\includegraphics[width=\columnwidth]{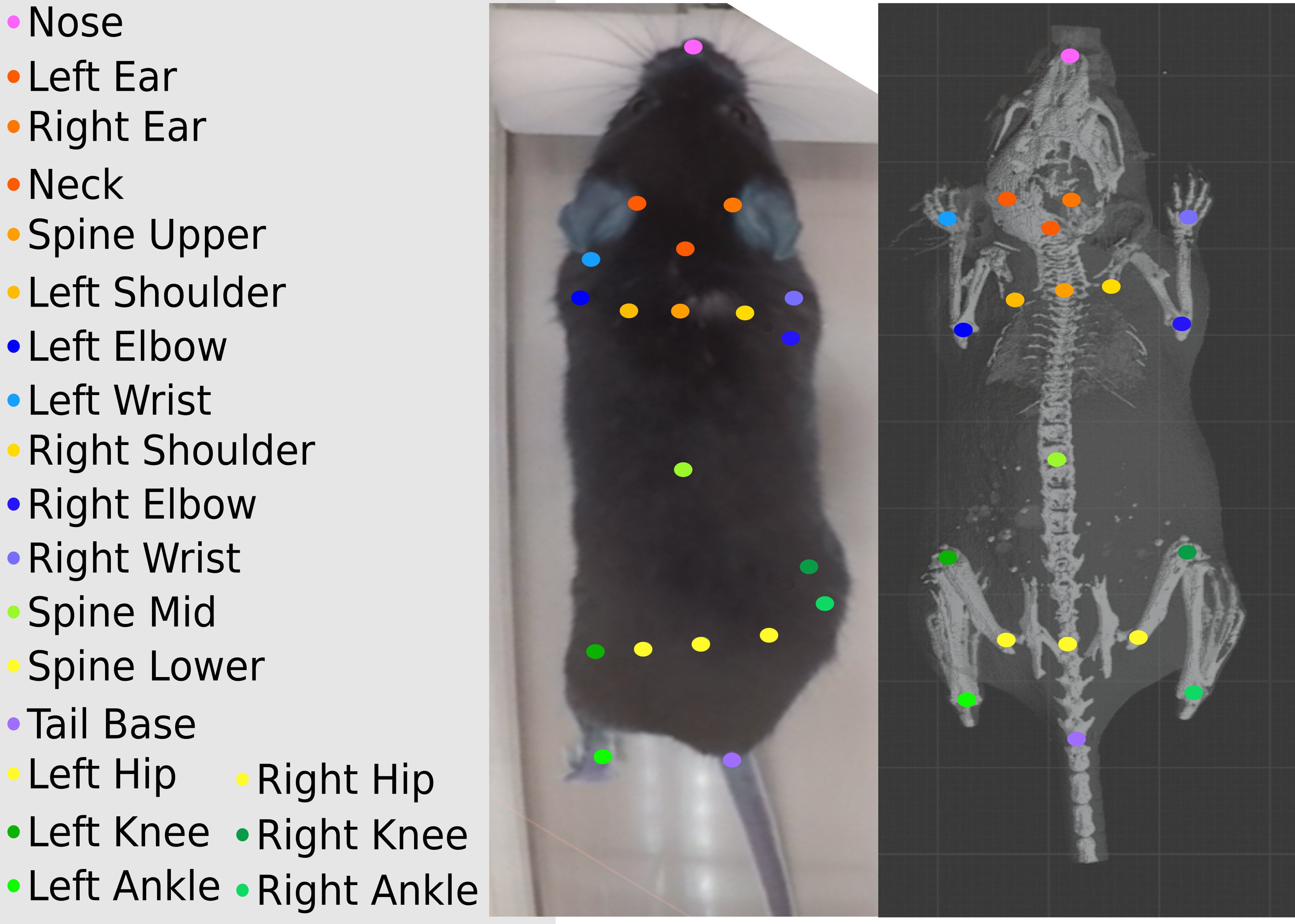}
\centering
\caption{\textit{left:} The labeled 2D keypoints. \textit{center:} A labeled image of a mouse with the joint legends to the left. \textit{right:} The high resolution CT scan segmented for bone in light colors, and segmented for the skin in darker colors with the corresponding keypoint locations at a neutral pose.}
\label{keypoint_annotation_fig}
\end{figure}

\subsubsection{3D pose prediction}
We adapt the human 3D pose optimization strategy from~\cite{bregler1998tracking} to mice because similar optimization strategies are successful with inferred 2D poses and relatively little 3D ground truth data~\cite{bogo2016keep}. We iteratively update the 3D joint angles on a kinematic chain consisting 18 joints corresponding to the 2D keypoints (the ears are excluded) to minimize the distance between the input 2D keypoint locations and the projected 3D joint locations. 

We improve the stability and convergence of the 3D pose optimization by using shape and/or pose priors~\cite{bogo2016keep}.
Specifically, we use a combination of joint angle constraints (joint angles must be within $\pm50$ degrees) and a Guassian Mixture Model pose prior constructed from a multiview reconstruction of the 3D pose (see below) and hand-posed models.
The hand-posed models have joint angles set in a 3D modeling software to match the apparent mouse pose in a set of images that cover typical poses. From these 3D poses, we align and scale the poses so that the vector from the base of the neck to the middle of the spine is defined as the x-axis and unit length, and then we fit a Gaussian mixture model with 5 components to the data. We optimize the joint angles to jointly minimized the reproduction error and maximize the likelihood under the mixture model.

The optimization is over-parameterized where the overall size and the distance to the camera are confounded, which can result in arbitrary scale and physically implausible rotations. We solve the complication by constraining the animal to a fixed distance from the camera. Similar scene constraints are a common approach to reconstructing physically meaningful 3D poses~\cite{arnab2019exploiting,zanfir2018monocular}.

\subsection{Multiview 3D Pose Reconstruction}
To generate ground truth 3D pose data for validation and constructing a pose prior, we build a custom, multiview 3D capture rig. A top-down RGB+D camera (Kinect) and two side RGB cameras with synchronized timing are calibrated with overlapping fields of view of a mouse cage. We label the 2D joint positions in synchronized frames from each field of view and triangulate the 3D location of each joint position that minimizes the reprojection errors. The multiview reconstructions are used to evaluate the single-view reconstruction quality and then added to the pose prior.  

\subsection{Biological Attribute Prediction}
To assess which representations preserve information about motion dynamics, we train a black-box model to predict biological attributes in the Continuous video data.
Because we want to study gait and not other factors, we limit the analysis to sequences when the animal is on or near the wheel during the night cycle when the mice are more active. We train on and predict labels for 10 second intervals, but evaluate performance across the aggregated prediction scores for each animal to normalize for the amount of time on the wheel. Data are split into the training and test sets with disjoint sets of mice in each. For each data representation we test, we train a convolutional neural network with kernel size 24 to predict each label independently. We perform a hyperparameter sweep over the number of layers in the network [2, 3, or 4], the number of hidden units in each layer [32, 64, 128, 256], and the learning rate [0.0001, 0.00001, 0.000001] using half the training set for validation. We report the best accuracy for each representation on the test set.

\subsection{Gait Measurements}
Direct measurements of gait parameters are obtained via a commercial system (DigiGait). We use the aggregated stride length from the Posture Plot report as well as the individual stride length measurements from the commercial system. We calculate similar measurements from our method by computing the duration of strides from the reconstructed pose and multiplying by the known treadmill speed to calculate the stride length. The aggregate duration of the stride is calculated as the wavelength of the Fourier spectrum peak magnitude and the individual stride durations are calculated as peak-to-peak times.
\section{Results}

\subsection{Inferred 3D Poses}

\begin{figure}
\includegraphics[width=\columnwidth]{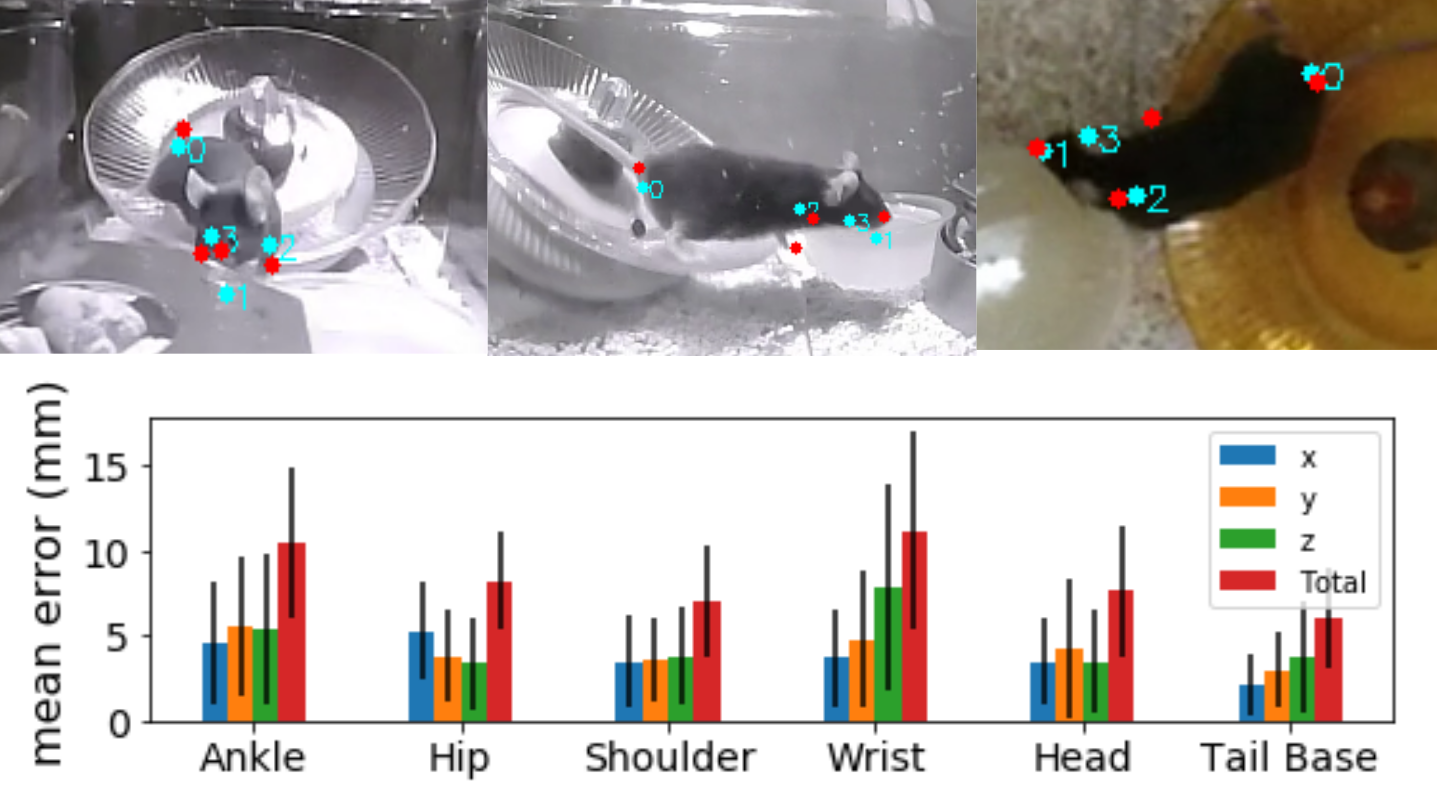}

\caption{Comparison of multi-view and single-view reconstructions.  The top three panels show three views of the mouse at the same time point. Red dots are reconstructions from triangulation and cyan dots from our single-view reconstruction. Four joints are shown as examples (0: tail, 1: noise, 2: left paw and 3: right paw).}
\label{fig:multiview-err}
\end{figure}

We quantitatively evaluate the quality of our 3D poses on the Multiview video data set. After determining the ground truth 3D pose from multiple views (see Methods~\ref{sec:methods}), we calculate how well we reconstruct the pose from the top down view alone. The inferred 3D pose is registered to the ground truth pose and we quantify the error in the inferred 3D pose in millimeters in Table~\ref{fig:multiview-err}. The average error for each joint is less than 10 mm. For common mouse body lengths of 10 cm, this represents less than 10\% relative error. We cannot find another monocular 3D pose reference that lists numbers to compare against. Although these numbers allow room for improvement, we demonstrate further results that this accuracy is sufficient to enable health predictions and extraction of gait parameters.

\subsection{Biological Attribute Prediction with 3D Pose}

After inferring the 3D poses, we address whether they add value to understanding biology. We use Continuous video data attributes---age, background, and knockout status---to assess how easily models can predict biological attributes from different features: the 2D bounding box, the 2D keypoints, the 3D keypoints, and the 3D joint angles. We train a range of artificial neural networks on each representation and present the best results for each feature on a held out set of 16 animals in Table~\ref{tab:bio-pred}. Of these, the 3D joint angles outperform the others by being able to perfectly classify each animal in the test set, while the others make one to three mistakes on the 16 test set animals.

\begin{table}[h]
    \centering
    \begin{tabular}{|c|c|c|c|}
        \hline
        Feature & acc(Age) & acc(Bkgrd) & acc(KO) \\
        \hline
        2D box & 0.86 $\pm$ 0.04 & 0.82 $\pm$ 0.01 & 0.89 $\pm$ 0.03 \\
        2D points & 0.85 $\pm$ 0.02 & 0.81 $\pm$ 0.02 & 0.91 $\pm$ 0.02 \\
        3D points & 0.88 $\pm$ 0.00 & 0.82 $\pm$ 0.02 & 0.90 $\pm$ 0.03 \\
        3D angles & \textbf{1.00 $\pm$ 0.00} & \textbf{1.00 $\pm$ 0.00} & \textbf{1.00 $\pm$ 0.00} \\
        \hline
    \end{tabular}
    \caption{Table of the classification accuracy (mean $\pm$ standard error of 5 training runs) for each input representation on a held-out set of animals for three attributes: Whether the animal is 12 or 3 months old (Age), whether the animals is a litter mate of a knockout (Bkgrd), and whether the animal is a knockout (KO). Best result in each column is in bold.}
    \label{tab:bio-pred}
\end{table}

\subsection{Accurate Gait Measurements from 3D Pose}

\begin{figure}[h!]
\includegraphics[width=\columnwidth]{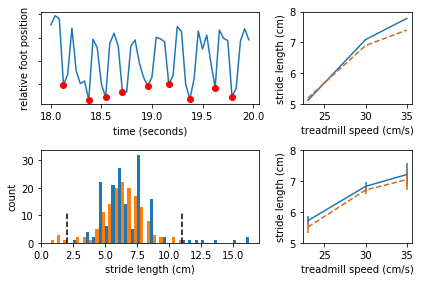}
\centering
\caption{\textit{top left:} An example time series of the foot position in arbitrary units. The periodic structure of gait is clearly visible. Red dots indicate peaks used in computing the stride length. \textit{top right:} The peak frequency in the foot position reconstruction $\times$ belt speed (blue, solid) and DigiGait posture plot stride length (orange, dashed). \textit{bottom left:} The distribution of stride lengths from the pose reconstruction (dark blue) and DigiGait (light orange). Dashed, black, vertical lines indicate outlier thresholds for statistical modeling. \textit{bottom right:} Stride lengths by treadmill speed for reconstructed pose (blue, solid) and DigiGait (orange, dashed). Error bars indicate $\pm$1 SEM.}
\label{fig:digigait}
\end{figure}

To further validate our method, we compare the strides measured by our system with the strides determined by the DigiGait system that directly images the feet. We infer the 3D poses as viewed from above using our method, estimate the strides and compare the output to the direct stride measurements by the DigiGait system in Figure~\ref{fig:digigait}. We find that we can recapitulate multiple direct measurements. 

The stride length estimated from the magnitude of the Fourier spectrum of the foot position over several seconds matches the aggregated Posture Plot stride length very well. Because the spectrum analysis aggregates over time, it should be more accurate than single stride analyses and avoids sampling noise due to the limited frame rate we use (24 fps). However, we cannot compute statistics from an aggregated number, so we also compared noisier individual stride estimates. 

We measure the peak-to-peak times to estimate the individual stride lengths and compare the distribution to the direct measurements. Excluding 13 asymmetric outliers beyond 2.3 $\sigma$ from the mean, the measurements from our system were not significantly different from the direct measurements (2-way ANOVA, main effect of measurement system: df=289, t=-0.8, p=0.424). While statistics cannot prove distributions are identical, we can claim that our measurements are similar to the commercial system.

\section{Conclusions}

Our method infers the 3D pose of mice from single view videos. This offers compelling opportunities for continuous, non-invasive monitoring. We demonstrate that the 3D joint angles enable predicting health related attributes of mice more easily than other features. Our system can even replace a custom hardware solution in determining gait parameters such as stride length. Future work includes improving the accuracy of the 3D pose and extending this method to animal social interactions.
{\small
\bibliographystyle{ieee_fullname}
\bibliography{egbib}

\begin{thebibliography}{10}\itemsep=-1pt

\bibitem{arnab2019exploiting}
Anurag Arnab, Carl Doersch, and Andrew Zisserman.
\newblock Exploiting temporal context for 3d human pose estimation in the wild.
\newblock In {\em Proceedings of the IEEE Conference on Computer Vision and
  Pattern Recognition}, pages 3395--3404, 2019.

\bibitem{bains2018assessing}
Rasneer~S Bains, Sara Wells, Rowland~R Sillito, J~Douglas Armstrong, Heather~L
  Cater, Gareth Banks, and Patrick~M Nolan.
\newblock Assessing mouse behaviour throughout the light/dark cycle using
  automated in-cage analysis tools.
\newblock {\em Journal of neuroscience methods}, 300:37--47, 2018.

\bibitem{bogo2016keep}
Federica Bogo, Angjoo Kanazawa, Christoph Lassner, Peter Gehler, Javier Romero,
  and Michael~J Black.
\newblock Keep it smpl: Automatic estimation of 3d human pose and shape from a
  single image.
\newblock In {\em European Conference on Computer Vision}, pages 561--578.
  Springer, 2016.

\bibitem{bregler1998tracking}
Christoph Bregler and Jitendra Malik.
\newblock Tracking people with twists and exponential maps.
\newblock In {\em Proceedings. 1998 IEEE Computer Society Conference on
  Computer Vision and Pattern Recognition (Cat. No. 98CB36231)}, pages 8--15.
  IEEE, 1998.

\bibitem{burn2013oxford}
David Burn.
\newblock {\em Oxford textbook of movement disorders}.
\newblock Oxford University Press, 2013.

\bibitem{cao2018openpose}
Zhe Cao, Gines Hidalgo, Tomas Simon, Shih-En Wei, and Yaser Sheikh.
\newblock Open{P}ose: realtime multi-person 2{D} pose estimation using {P}art
  {A}ffinity {F}ields.
\newblock In {\em arXiv preprint arXiv:1812.08008}, 2018.

\bibitem{deacon2013measuring}
Robert~MJ Deacon.
\newblock Measuring motor coordination in mice.
\newblock {\em JoVE (Journal of Visualized Experiments)}, (75):e2609, 2013.

\bibitem{dorman2014comparison}
Christopher~W Dorman, Hollis~E Krug, Sandra~P Frizelle, Sonia Funkenbusch, and
  Maren~L Mahowald.
\newblock A comparison of digigait™ and treadscan™ imaging systems:
  assessment of pain using gait analysis in murine monoarthritis.
\newblock {\em Journal of pain research}, 7:25, 2014.

\bibitem{gould2009open}
Todd~D Gould, David~T Dao, and Colleen~E Kovacsics.
\newblock The open field test.
\newblock In {\em Mood and anxiety related phenotypes in mice}, pages 1--20.
  Springer, 2009.

\bibitem{hong2015automated}
Weizhe Hong, Ann Kennedy, Xavier~P Burgos-Artizzu, Moriel Zelikowsky,
  Santiago~G Navonne, Pietro Perona, and David~J Anderson.
\newblock Automated measurement of mouse social behaviors using depth sensing,
  video tracking, and machine learning.
\newblock {\em Proceedings of the National Academy of Sciences},
  112(38):E5351--E5360, 2015.

\bibitem{jhuang2010automated}
Hueihan Jhuang, Estibaliz Garrote, Xinlin Yu, Vinita Khilnani, Tomaso Poggio,
  Andrew~D Steele, and Thomas Serre.
\newblock Automated home-cage behavioural phenotyping of mice.
\newblock {\em Nature communications}, 1(1):1--10, 2010.

\bibitem{johnson2016composing}
Matthew~J Johnson, David~K Duvenaud, Alex Wiltschko, Ryan~P Adams, and
  Sandeep~R Datta.
\newblock Composing graphical models with neural networks for structured
  representations and fast inference.
\newblock In {\em Advances in neural information processing systems}, pages
  2946--2954, 2016.

\bibitem{kabra2013jaaba}
Mayank Kabra, Alice~A Robie, Marta Rivera-Alba, Steven Branson, and Kristin
  Branson.
\newblock Jaaba: interactive machine learning for automatic annotation of
  animal behavior.
\newblock {\em Nature methods}, 10(1):64, 2013.

\bibitem{humanMotionKZFM19}
Angjoo Kanazawa, Jason~Y. Zhang, Panna Felsen, and Jitendra Malik.
\newblock Learning 3d human dynamics from video.
\newblock In {\em Computer Vision and Pattern Recognition (CVPR)}, 2019.

\bibitem{lassner2017unite}
Christoph Lassner, Javier Romero, Martin Kiefel, Federica Bogo, Michael~J
  Black, and Peter~V Gehler.
\newblock Unite the people: Closing the loop between 3d and 2d human
  representations.
\newblock In {\em Proceedings of the IEEE Conference on Computer Vision and
  Pattern Recognition}, pages 6050--6059, 2017.

\bibitem{lin2014coco}
Tsung-Yi Lin, Michael Maire, Serge Belongie, James Hays, Pietro Perona, Deva
  Ramanan, Piotr Doll{\'a}r, and C~Lawrence Zitnick.
\newblock Microsoft coco: Common objects in context.
\newblock In {\em European conference on computer vision}, pages 740--755.
  Springer, 2014.

\bibitem{liu2016ssd}
Wei Liu, Dragomir Anguelov, Dumitru Erhan, Christian Szegedy, Scott Reed,
  Cheng-Yang Fu, and Alexander~C Berg.
\newblock Ssd: Single shot multibox detector.
\newblock In {\em European conference on computer vision}, pages 21--37.
  Springer, 2016.

\bibitem{liu2019towards}
Zhenguang Liu, Shuang Wu, Shuyuan Jin, Qi Liu, Shijian Lu, Roger Zimmermann,
  and Li Cheng.
\newblock Towards natural and accurate future motion prediction of humans and
  animals.
\newblock In {\em Proceedings of the IEEE Conference on Computer Vision and
  Pattern Recognition}, pages 10004--10012, 2019.

\bibitem{loper2015smpl}
Matthew Loper, Naureen Mahmood, Javier Romero, Gerard Pons-Moll, and Michael~J
  Black.
\newblock Smpl: A skinned multi-person linear model.
\newblock {\em ACM transactions on graphics (TOG)}, 34(6):248, 2015.

\bibitem{mathis2018deeplabcut}
Alexander Mathis, Pranav Mamidanna, Kevin~M Cury, Taiga Abe, Venkatesh~N
  Murthy, Mackenzie~Weygandt Mathis, and Matthias Bethge.
\newblock Deeplabcut: markerless pose estimation of user-defined body parts
  with deep learning.
\newblock {\em Nature neuroscience}, 21(9):1281, 2018.

\bibitem{newell2016stacked}
Alejandro Newell, Kaiyu Yang, and Jia Deng.
\newblock Stacked hourglass networks for human pose estimation.
\newblock In {\em European conference on computer vision}, pages 483--499.
  Springer, 2016.

\bibitem{noldus2001ethovision}
Lucas~PJJ Noldus, Andrew~J Spink, and Ruud~AJ Tegelenbosch.
\newblock Ethovision: a versatile video tracking system for automation of
  behavioral experiments.
\newblock {\em Behavior Research Methods, Instruments, \& Computers},
  33(3):398--414, 2001.

\bibitem{pavlakos2018learning}
Georgios Pavlakos, Luyang Zhu, Xiaowei Zhou, and Kostas Daniilidis.
\newblock Learning to estimate 3d human pose and shape from a single color
  image.
\newblock In {\em Proceedings of the IEEE Conference on Computer Vision and
  Pattern Recognition}, pages 459--468, 2018.

\bibitem{pereira2019fast}
Talmo~D Pereira, Diego~E Aldarondo, Lindsay Willmore, Mikhail Kislin, Samuel
  S-H Wang, Mala Murthy, and Joshua~W Shaevitz.
\newblock Fast animal pose estimation using deep neural networks.
\newblock {\em Nature methods}, 16(1):117--125, 2019.

\bibitem{richardson2015power}
Claire~A Richardson.
\newblock The power of automated behavioural homecage technologies in
  characterizing disease progression in laboratory mice: A review.
\newblock {\em Applied Animal Behaviour Science}, 163:19--27, 2015.

\bibitem{salem2019three}
Ghadi Salem, Jonathan Krynitsky, Monson Hayes, Thomas Pohida, and Xavier
  Burgos-Artizzu.
\newblock Three-dimensional pose estimation for laboratory mouse from monocular
  images.
\newblock {\em IEEE Transactions on Image Processing}, 28(9):4273--4287, 2019.

\bibitem{sheets2013quantitative}
Alison~L Sheets, Po-Lun Lai, Lesley~C Fisher, and D~Michele Basso.
\newblock Quantitative evaluation of 3d mouse behaviors and motor function in
  the open-field after spinal cord injury using markerless motion tracking.
\newblock {\em PloS one}, 8(9), 2013.

\bibitem{tung2017self}
Hsiao-Yu Tung, Hsiao-Wei Tung, Ersin Yumer, and Katerina Fragkiadaki.
\newblock Self-supervised learning of motion capture.
\newblock In {\em Advances in Neural Information Processing Systems}, pages
  5236--5246, 2017.

\bibitem{varol2017learning}
Gul Varol, Javier Romero, Xavier Martin, Naureen Mahmood, Michael~J Black, Ivan
  Laptev, and Cordelia Schmid.
\newblock Learning from synthetic humans.
\newblock In {\em Proceedings of the IEEE Conference on Computer Vision and
  Pattern Recognition}, pages 109--117, 2017.

\bibitem{wei2016convolutional}
Shih-En Wei, Varun Ramakrishna, Takeo Kanade, and Yaser Sheikh.
\newblock Convolutional pose machines.
\newblock In {\em Proceedings of the IEEE conference on Computer Vision and
  Pattern Recognition}, pages 4724--4732, 2016.

\bibitem{wiltschko2015mapping}
Alexander~B Wiltschko, Matthew~J Johnson, Giuliano Iurilli, Ralph~E Peterson,
  Jesse~M Katon, Stan~L Pashkovski, Victoria~E Abraira, Ryan~P Adams, and
  Sandeep~Robert Datta.
\newblock Mapping sub-second structure in mouse behavior.
\newblock {\em Neuron}, 88(6):1121--1135, 2015.

\bibitem{xu2019gait}
Yu Xu, Na-Xi Tian, Qing-Yang Bai, Qi Chen, Xiao-Hong Sun, and Yun Wang.
\newblock Gait assessment of pain and analgesics: comparison of the digigait™
  and catwalk™ gait imaging systems.
\newblock {\em Neuroscience bulletin}, 35(3):401--418, 2019.

\bibitem{zanfir2018monocular}
Andrei Zanfir, Elisabeta Marinoiu, and Cristian Sminchisescu.
\newblock Monocular 3d pose and shape estimation of multiple people in natural
  scenes-the importance of multiple scene constraints.
\newblock In {\em Proceedings of the IEEE Conference on Computer Vision and
  Pattern Recognition}, pages 2148--2157, 2018.

\end{thebibliography}
}

\end{document}